\RequirePackage[l2tabu, orthodox]{nag}
\documentclass[11pt]{article}

\usepackage[T1]{fontenc}

\usepackage{soul} 

\usepackage{garamond}  
\renewcommand{\rmdefault}{ugm}

\usepackage[bitstream-charter]{mathdesign}
\usepackage{amsmath}
\usepackage[scaled=0.92]{PTSans}

\usepackage[
  paper  = letterpaper,
  left   = 1.0in,
  right  = 1.0in,
  top    = 1.0in,
  bottom = 1.0in,
  ]{geometry}

\usepackage[usenames,dvipsnames]{xcolor}
\definecolor{shadecolor}{gray}{0.9}

\usepackage[final,expansion=alltext]{microtype}
\usepackage[english]{babel}
\usepackage[parfill]{parskip}
\usepackage{afterpage}
\usepackage{framed}

{\endMakeFramed}

\usepackage{lineno}

\usepackage{ragged2e}

\newcounter{parcount}

\usepackage{graphicx}
\usepackage[labelfont=bf]{caption}

\usepackage{booktabs}
\usepackage{multirow}

\usepackage[algoruled]{algorithm2e}
\usepackage{listings}
\usepackage{fancyvrb}
\fvset{fontsize=\normalsize}

\usepackage{natbib}

\usepackage[acronym,nowarn]{glossaries}
\makeglossaries

\definecolor{peru}{rgb}{0.8, 0.52, 0.25}
\usepackage[colorlinks,linktoc=all]{hyperref}
\usepackage[all]{hypcap}
\hypersetup{citecolor=peru}
\hypersetup{linkcolor=peru}
\hypersetup{urlcolor=peru}

\usepackage[nameinlink]{cleveref}
\creflabelformat{equation}{#1#2#3}
\crefname{equation}{eq.}{eqs.}  
\Crefname{equation}{Eq.}{Eqs.}

\lstdefinestyle{mystyle}{
    commentstyle=\color{OliveGreen},
    keywordstyle=\color{BurntOrange},
    numberstyle=\tiny\color{black!60},
    stringstyle=\color{MidnightBlue},
    basicstyle=\ttfamily,
    breakatwhitespace=false,
    breaklines=true,
    captionpos=b,
    keepspaces=true,
    numbers=left,
    numbersep=5pt,
    showspaces=false,
    showstringspaces=false,
    showtabs=false,
    tabsize=2
}
\usepackage[colorinlistoftodos,
           prependcaption,
           textsize=small,
           backgroundcolor=yellow,
           linecolor=lightgray,
           bordercolor=lightgray]{todonotes}

\usepackage{amsthm}  

\usepackage{bm}

\usepackage[colorinlistoftodos,
           prependcaption,
           textsize=small,
           backgroundcolor=yellow,
           linecolor=lightgray,
           bordercolor=lightgray]{todonotes}

\usepackage{graphicx}
\usepackage{caption}
\usepackage{subcaption}
\usepackage{wrapfig}

\usepackage{booktabs}
\usepackage{arydshln} 
\usepackage{multirow}
\usepackage{nicematrix}

\usepackage{listings}
\usepackage{fancyvrb}
\fvset{fontsize=\normalsize}

\usepackage[nameinlink]{cleveref}
\creflabelformat{equation}{#1#2#3}
\crefname{equation}{eq.}{eqs.}  
\Crefname{equation}{Eq.}{Eqs.}

\usepackage{listings}
\lstdefinestyle{alp_style}{
    commentstyle=\color{OliveGreen},
    numberstyle=\tiny\color{black!60},
    stringstyle=\color{BrickRed},
    basicstyle=\ttfamily\scriptsize,
    breakatwhitespace=false,
    breaklines=true,
    captionpos=b,
    keepspaces=true,
    numbers=none,
    numbersep=5pt,
    showspaces=false,
    showstringspaces=false,
    showtabs=false,
    tabsize=2
}

\usepackage{capt-of}

\usepackage{algorithmicx}

\usepackage{lipsum}

\newtheorem{proposition}{Proposition}[section]

\theoremstyle{remark}
\newtheorem*{lemma*}{Lemma}

\newcommand{\bX}{\boldsymbol{X}}
\newcommand{\bK}{\boldsymbol{K}}

\newcommand{\tTr}{\operatorname{Tr}}

\newcommand{\rhog}{\rho_{\text{global}}}

\usepackage{authblk}
\usepackage{enumitem} 
\usepackage{tabularx}
\usepackage{booktabs}
\usepackage{pifont}
\usepackage[T1]{fontenc}
\usepackage{multirow}
\usepackage{threeparttable}

\definecolor{best_color}{RGB}{200,120,0}  
\definecolor{second_color}{RGB}{80,125,190}  
\definecolor{third_color}{RGB}{180,158,214}
\newcommand{\best}[1]{\textbf{\textcolor{best_color}{#1}}}
\newcommand{\second}[1]{\textbf{\underline{\textcolor{second_color}{#1}}}}
\newcommand{\third}[1]{\textbf{\underline{\textcolor{third_color}{#1}}}}

\usepackage[most]{tcolorbox}

\tcolorboxenvironment{proposition}{
  colback=gray!10,
  colframe=gray!60,
  boxrule=0.5pt,
  arc=3pt,
  left=6pt,
  right=6pt,
  top=6pt,
  bottom=6pt
}

\title{\textbf{Vendi Novelty Scores for Out-of-Distribution Detection}}

\author[1, 3]{Amey Pasarkar}
\author[2, 3]{Adji Bousso Dieng}
\affil[1]{Lewis-Sigler Institute For Integrative Genomics, Princeton University}
\affil[2]{Department of Computer Science, Princeton University}
\affil[3]{\href{https://vertaix.princeton.edu/}{Vertaix}}

\begin{document}
\maketitle

\begin{abstract}
\noindent Out-of-distribution (OOD) detection is critical for the safe deployment of machine learning systems and for scientific discovery. Existing post-hoc detectors typically rely on model confidence scores or likelihood estimates in feature space, often under restrictive distributional assumptions. In this work, we introduce a new paradigm and formulate OOD detection from a diversity perspective. We propose the Vendi Novelty Score (VNS), an OOD detector based on the Vendi Scores (VS), a family of similarity-based diversity metrics. VNS quantifies how much a test sample increases the VS of the in-distribution feature set, providing a principled notion of novelty that does not require density modeling. VNS is linear-time and naturally combines class-conditional (local) and dataset-level (global) novelty signals. Across multiple image classification benchmarks and network architectures, VNS achieves state-of-the-art OOD detection performance. Remarkably, VNS retains this performance when computed using only 1\% of the training data, enabling deployment in memory- or access-constrained settings.\\

\end{abstract}

\section{Introduction}
Modern machine learning systems have made remarkable progress across a wide variety of tasks. Yet, these systems can behave unpredictably on previously unseen out-of-distribution (OOD) inputs, often producing high-confidence, incorrect predictions \citep{goodfellow2014explaining, hendrycks2016baseline}. Knowing when a model can be trusted is essential in high-stakes settings such as autonomous driving \citep{shoeb2025out}, medical imaging \citep{zhang2021out}, and model-guided scientific discovery \citep{segal2025known}. OOD detectors provide a safeguard for these settings by identifying the inputs that are OOD and should be routed to human review or other fallback policies.

Recent methods for OOD detection span specialized models \citep{hendrycks2018deep, katz2022training}, modifications to the training process \citep{pinto2022using, hendrycks2022pixmix}, and post-hoc methods that act independently of the training process \citep{hendrycks2016baseline, lee2018simple,  yang2024generalized}. As models have become increasingly large and trained on massive datasets, post-hoc methods are often the cheapest and most practical options for OOD detection. Post-hoc methods can be broadly categorized by the information from the model they exploit during inference. One class of approaches relies on the model's prediction logits \citep{liu2023gen, liu2020energy}, while another leverages the model's intermediate representations and gradients to identify deviations from the in-distribution (ID) feature geometry \citep{ren2021simple, ammar2023neco, huang2021importance}. While this second class of methods often achieves the strongest performance, they also make strong assumptions about the geometry of the model's representations, which may not hold consistently across architectures and datasets. For example, \citet{liu2025detecting} leverages the properties of neural collapse for OOD detection, which can lead to worse performance on off-the-shelf classifiers that do not exhibit neural collapse. Other similarity-based approaches like \citet{sun2022out} do not make explicit assumptions about the representation space, but incur significant test-time overhead by performing a k-nearest neighbor search to identify OOD samples. More recent work aiming to exploit the gradients of the model \citep{regmi2025adascale} requires multiple forward passes, making them impractical in latency-sensitive settings. Designing computationally efficient OOD detectors without restrictive distributional assumptions is therefore critical for real-world deployment.

\begin{figure*}
    \centering
    \includegraphics[width=0.8\linewidth,
    trim=2.5cm 7.3cm 10cm 9.5cm, clip]{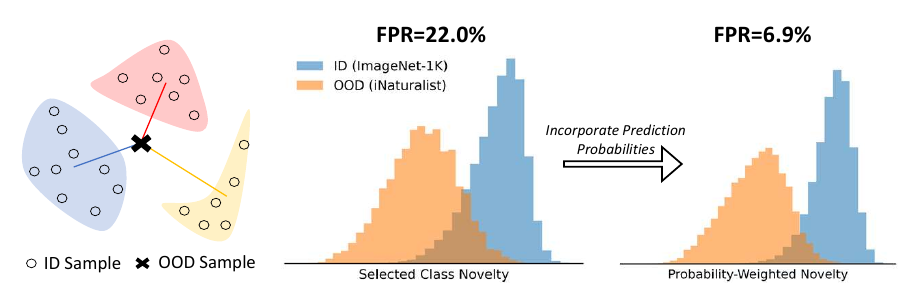}
    \caption{Conceptual overview of VNS. Left: Three in-distribution (ID) classes are shown as clusters in representation space (colored regions), along with a test out-of-distribution (OOD) sample (star). VNS computes a class-conditional novelty score for the OOD sample. Middle: Using only the class-conditional novelty score from the model's predicted class provides reasonable ID-OOD separation. Right: Aggregating class-conditional novelty scores using prediction probabilities (Equation \ref{eq:aggregate}) yields improved ID-OOD separation, reducing the False Positive Rate (FPR).}
    \label{fig:schematic}
\end{figure*}

In this work, we introduce a new paradigm for post-hoc OOD detection based on diversity. We build on the Vendi Scores (VS), a family of similarity-based diversity metrics \citep{friedman2022vendi, pasarkar2023cousins}, and propose the Vendi Novelty Score (VNS). Rather than approximating feature distributions, VNS quantifies novelty via a test sample's impact on representation diversity. Specifically, VNS computes class-conditional novelty contributions---as measured by the change in VS when the test sample is added to each training set class---and aggregates these contributions using the model’s predicted class probabilities. Figure~\ref{fig:schematic} provides an illustration.

VNS offers an efficient way to evaluate sample novelty and incorporates both local and global novelty signals. We provide a series of approximations that allow VNS to scale linearly with the data dimension and number of classes. We demonstrate the effectiveness of VNS across various OOD benchmarks spanning several datasets (CIFAR-10, CIFAR-100 \citep{krizhevsky2009learning}, ImageNet-1K \citep{deng2009imagenet}) and multiple network architectures (ResNet \citep{he2016deep}, ViT \citep{dosovitskiy2020image}, Swin-T \citep{liu2021swin}). Our approach often achieves state-of-the-art detection performance, even when given access to only 1\% of the training data. We also provide insights into the accuracy and efficiency of VNS theoretically and through various ablation studies. 

\section{Related Work}
\label{sec:related}
\textbf{Post-Hoc OOD Detection.}
Early work on post-hoc OOD detection focused on score functions based on the model's logits. Such methods leverage characteristics of the logit distribution, such as the maximum class probabilities, energy scores, or their entropy \citep{hendrycks2016baseline, liu2020energy, liu2023gen}. \citet{huang2021importance} measured novelty via the gradient norm of the KL divergence of the logit distribution from a uniform prior. These approaches are attractive because they do not rely on access to the training data, making them easy to deploy. However, this simplicity comes at the cost of ignoring in-distribution feature statistics, which can reduce OOD detection reliability.

Another line of work modifies the inference-time activations to increase separability between ID and OOD logit distributions. Activation-shaping methods clip extreme activations or apply pruning or adaptive scaling to activations \citep{sun2021react, djurisic2022extremely, xu2023scaling}. Input perturbation methods such as ODIN \citep{liang2017enhancing}, IODIN \citep{regmi2024image}, and ADASCALE \citep{regmi2025adascale} use gradient information to modify the input at test time, but require additional forward and backward passes. 

Most closely related to our approach are feature-space density methods based on Mahalanobis distance. These methods model features as class-conditional Gaussians \citep{lee2018simple} with a background global Gaussian score \citep{ren2021simple}. \citet{mueller2025mahalanobis++} found that normalizing samples prior to fitting the Gaussian densities helps improve detection robustness across architectures. Related feature-space methods such as ViM \citep{wang2022vim} combine information from logits and intermediate representations, modeling ID structure via a principal subspace of features and using the residual norm to improve OOD separation.
Other geometry-based methods include FDBD---which found that ID samples reside closer to decision boundaries than OOD samples \citep{liu2023fast}---and methods inspired by neural collapse \citep{ammar2023neco, liu2025detecting}. These approaches are significantly faster than the k-nearest neighbor approach proposed in \citet{sun2022out}.

\textbf{Training Methods for OOD Detection.}
A complementary approach for OOD detection is to modify the training process of the model so that OOD samples are easier to detect. Researchers have explored modifying the architecture itself, like in \citet{devries2018learning}, where a separate confidence-based OOD detector branch was added to the model. Alternatively, many works have explored modifications to the training data. For example, adding OOD samples and augmentations has been shown to improve detection accuracy \citep{zhang2023matrix, zhang2023mixture, zhu2023diversified, pinto2022using}. Other methods have explored new training objectives, such as tuning logit activations \citep{xu2023scaling,wei2022mitigating}, or using contrastive learning to improve ID and OOD separation \citep{seifi2024ood, winkens2020contrastive, tack2020csi, ming2022exploit}. VNS is complementary to these methods, since it operates post-hoc and can be applied to pretrained models without retraining. 

\textbf{Vendi Scoring for Novelty.}
The Vendi Score, introduced in \citep{friedman2022vendi}, has been used as an accurate measure of diversity across domains. Prior work has used the Vendi Score to evaluate dataset diversity in machine learning and the natural sciences \citep{nielsen2025applications, dieng2025unified, roediger2025machine, priyadarshini2025diversity, bai2025dragon}, as well as to encourage diverse outputs in simulations, generative models, and reinforcement learning \citep{rezaei2025alpha, rezaei2025vendi, pasarkar2023vendi, askari2024improving, lintunen2025vendirl}. Recent work has also explored Vendi-based novelty criteria.
For example, \citet{nguyen2024quality} and \citet{liu2024diversity} measure the incremental change in Vendi Score induced by adding a candidate point as a selection rule in active learning. More recently, \citet{pasarkar2025vendiscope} proposed the Vendiscope, a method to analyze the content of large-scale datasets. The Vendiscope quantifies sample rarity by optimizing over a distribution supported on the dataset points. This formulation cannot measure an unseen sample's rarity \textit{against} a dataset, so it is not applicable to OOD detection.

\section{The Vendi Novelty Score}
In this section, we briefly review the Vendi Scores (VS), and introduce our proposed algorithm to make them effective for OOD detection.

\textbf{The Vendi Scores.}
Consider a set of $N$ data points $\bX \in \mathbb{R}^{N \times D}$. 
Let $k(\cdot, \cdot)$ denote a positive semi-definite pairwise kernel. The VS, as defined in \citet{friedman2022vendi}, is the entropy of the eigenvalues of the similarity matrix $\bK \in \mathbb{R}^{N \times N}$ induced by the kernel $k$, normalized so that $\text{Tr}(\bK)=1$. This can be generalized to the \text{R\'enyi} entropy \citep{pasarkar2023cousins}
\begin{equation}
\label{eq:vs-general}
 \text{VS}_q(\bX, k) = \exp\left(\frac{1}{1-q}\log \sum_{i=1}^{N} \lambda_{i}(\bK)^q\right),
\end{equation}
where $\lambda_i(\bK)$ is the $i$th eigenvalue of $\bK$ and $q \geq 0$ is the order of the VS.

Computing the VS for arbitrary similarity matrices $\bK$ requires computing their eigenspectrum, which requires $O(N^3)$ time. However, in the common setting where each data point is represented by an $\ell_2$-normalized vector and the cosine kernel is used--as is often the case for learned embeddings--\citet{friedman2022vendi} notes that we can compute the VS more cheaply. In this case, the similarity matrix is $\bK = \bX \bX^T /N$, and its non-zero eigenvalues coincide with those of the matrix $\rho = \bX^T \bX /N \in \mathbb{R}^{D \times D}$. We can compute the VS using $\rho$, which reduces the time complexity of VS computation to $O(D^2 N + D^3)$ and is efficient when $N \gg D$. 

Certain choices of the order $q$ have convenient forms. We highlight $q=2$ and $q=\infty$: 
\begin{align}
\label{eq:trace}
    \text{VS}_2(\bX, k) = \frac{1}{\tTr(\bK^2)}  = \frac{1}{\sum_{i, j=1}^N K_{ij}^2} \text{ and } \text{VS}_\infty(\bX, k) = \frac{1}{\lambda_{\max} (\bK)}.
\end{align}

\textbf{Class-Conditional Novelty.} Here, we describe how we use the VS to measure the novelty of a test sample with respect to a class. We consider a trained discriminative model $f_\theta$ (e.g., ResNet-50) with $C$ classes, trained on a labeled in-distribution dataset $\mathcal{D}_{\text{train}} = \{(x_j, y_j)\}_{j=1}^N$. Let $N_c$ denote the number of training samples belonging to class $c$. Let $h(x) \in \mathbb{R}^D$ denote the representation of input $x$ extracted from the penultimate layer of $f_\theta$, and let $\bX_c \in \mathbb{R}^{N_c \times D}$ be the matrix of embeddings $h(x)$ of all samples that belong to class $c$ in $\mathcal{D}_{\text{train}}$. All representations are $\ell_2$-normalized before processing.

We first compute class-conditional matrices $\rho_{c}  = \bX_c^T \bX_c / N_c \in \mathbb{R}^{D \times D}$. Given a test sample $x$, we consider the class-conditional matrix obtained after incorporating the representation $h(x)$ into class $c$, which we denote by $\rho'_c$. We denote by $\text{VS}_2(\rho_c)$ the VS of order $2$ computed for class $c$ under a cosine kernel. The class-conditional novelty is then defined as
\begin{align}
    \Delta_c(x) = \log \text{VS}_2(\rho'_{c}) - \log \text{VS}_2(\rho_c). 
\end{align}
$\Delta_c(x)$ measures the log-ratio increase in diversity from adding sample $x$ to class $c$. 

We can use Equation \ref{eq:trace} to compute $\Delta_c$ efficiently, using the identity $\text{VS}_2(\rho_c) = \frac{1}{\tTr(\rho_c^2)}$. To do so, the updated matrix after adding a test sample $x$ can be written as $\rho'_c = \frac{N_c\rho_c + h(x) h(x)^T}{N_c+1}$. Substituting this expression yields 
\begin{align*}
    \tTr({\rho'_c}^2) &= \frac{N_c^2 \tTr(\rho_c^2) + 2N_c h(x)^T \rho_c h(x) + 1}{(N_c+1)^2}.
\end{align*}
Denote the eigenvalues and eigenvectors of $\rho_c$ as $\lambda_{c,1}, \lambda_{c,2},\ldots,\lambda_{c,D}$ and $u_{c,1}, u_{c,2},\ldots,u_{c,D}$. Let $\alpha_{c,j}(x) = (u_{c,j}^T h(x))^2$. We can rewrite the above equation for $\tTr({\rho'_c}^2)$ in terms of this eigenbasis, 
\begin{align}
    \tTr({\rho'_c}^2) = \frac{N_c^2 \sum_{i=1}^D\lambda_{c,i}^2 + 2N_c \sum_{i=1}^D \lambda_{c,i} \alpha_{c,i}(x) + 1}{(N_c+1)^2}.
\end{align}
Plugging in $\tTr(\rho_c'^2)$ into Equation \ref{eq:trace} would give us an exact, closed-form, $O(D^2)$ time complexity operation to compute the class-conditional novelty scores. However, we find that maintaining the entire eigenbasis is unnecessary and can be sensitive to estimation noise in small classes. Instead, we can achieve strong results if we only maintain the top eigenvalue (Appendix Section \ref{sec:hyperparam}). Let $\lambda_c$ denote the largest eigenvalue for class $c$, with corresponding eigenvector $u_c$.
Then, we can compute the new trace in $O(D)$ using the rank-1 approximation
\begin{align*}
    \alpha_c(x) := (u_c^T h(x)) ^2 \qquad 
    \tTr({\rho'_c}^2) \approx  \frac{N_c^2 \lambda_{c}^2 + 2N_c \lambda_{c}\alpha_{c}(x) + 1}{(N_c+1)^2}.
\end{align*}
The class-conditional novelty contribution is then
\begin{align}
\label{eq:class}
    \Delta_c(x) = -\log \left( \frac{N_c^2 \lambda_{c}^2 + 2N_c \lambda_{c}\alpha_{c}(x) + 1}{(N_c+1)^2} \right) + \log(\lambda_c^2).
\end{align}
For each test sample $x$, we compute a vector of class-conditional novelty scores
\begin{align}
\boldsymbol{\Delta}(x) = (\Delta_1(x), \Delta_2(x), \dots, \Delta_C(x)).
\end{align}

\textbf{Probability-weighted aggregation.}
The classifier $f_\theta$ produces a predictive distribution
\[
\mathbf{p}(x) = (p_1(x), \dots, p_C(x)),
\quad
p_c(x) = \Pr(y=c \mid x).
\]
Rather than relying on the novelty with respect to a single predicted class, we aggregate class-conditional novelty scores using a probability-weighted scheme. Let $\mathcal{T}_K(x)\subseteq\{1,\dots,C\}$ denote the indices of the $K$ classes with largest predicted probabilities $p_c(x)$.
\begin{align}
\label{eq:aggregate}
S_{\text{LOCAL-OOD}}(x)
=
\sum_{c \in \mathcal{T}_K(x)} N_c p_c(x)^{\gamma} \Delta_c(x),
\end{align}
where $\gamma \ge 0$ and $K \leq C$ are tunable hyperparameters. 

Notably, Eq.~\eqref{eq:aggregate} mirrors the form of GEN \citep{liu2023gen}, which aggregates over $\mathcal{T}_K(x)$ using probability-based weights of the form $p_c(x)^\gamma(1-p_c(x))^\gamma$; in contrast, we use $p_c(x)^\gamma$ to weight class-conditional novelty terms $\Delta_c(x)$. The exponent $\gamma$ controls the sharpness of the aggregation: larger values emphasize high-confidence classes, while smaller values yield a softer, more global aggregation across classes. The parameter $K$ controls the number of ID classes included in the aggregation (selected as the top-$K$ classes by $p_c(x)$). While increasing $K$ can incorporate information when predictions are ambiguous, overly large values may degrade performance by including many low-probability classes whose class-conditional novelty scores are weakly related to the sample and introduce noise into the aggregation. Smaller values of $K$ also enable cheaper computation.

We additionally rescale $\Delta_c(x)$ by $N_c$ to account for class-size. Appendix Section \ref{subsec:scaling} shows that the class-conditional novelty score scales by $O(1/N_c)$, so we remove the effect of class-size with this factor.. 

\textbf{Incorporating Global Diversity.}
Previous work has highlighted the importance of modeling the local density of model representations as well as the global density. \citet{ren2019likelihood} and \citet{ren2021simple} showed that subtracting a global novelty or likelihood score from the local novelty score can yield more robust results. Following this idea, we model a background density using a first-order approximation of the effect of a test sample $x$ on the $\text{VS}_\infty$ computed on the entire dataset.

Let $\bX \in \mathbb{R}^{N \times D}$ denote the matrix whose rows are the $\ell_2$-normalized embeddings
$h(x)$ of all training samples, with $\rhog = \bX^T \bX / N \in \mathbb{R}^{D \times D}$. Given a test sample $x$, we denote the updated global matrix as $\rhog'$ and define the global novelty as
\begin{align}
\label{eq:global_def}
    \Delta_{\text{global}}(x) = \log \text{VS}_\infty(\rhog') - \log \text{VS}_\infty(\rhog). 
\end{align}
From Equation \ref{eq:trace}, the $\text{VS}_\infty$ only depends on the maximum eigenvalue of $\rho_g$. We can approximate the effect of $x$ on the largest eigenvalue $\lambda_{\max}$ using Proposition \ref{prop:lambda_max_update}, with proof in Section \ref{subsec:rayleigh}.

\begin{proposition}[Accuracy of the Max Eigenvalue Update]
\label{prop:lambda_max_update}
Let $\rhog \in \mathbb{R}^{D\times D}$ be a symmetric positive semidefinite matrix with largest eigenvalue
$\lambda_{\max}(\rhog)$ and corresponding unit-norm eigenvector $u_{\max}$.
For a unit-norm vector $h(x)\in\mathbb{R}^D$ and dataset size $N$, define the rank-one updated matrix
\begin{align*}
    \rhog' = \frac{1}{N+1}\left(N\rhog + h(x)h(x)^T\right).
\end{align*}
Then the largest eigenvalue of $\rhog'$ admits the first-order expansion
\begin{align*}
\lambda_{\max}(\rhog') =&
\frac{N\lambda_{\max}(\rhog) + (u_{\max}^T h(x))^2}{N+1}  +O\!\left(\frac{1}{N^2}\right).
\end{align*}
Equivalently, the estimator
\begin{align*}
\widehat{\lambda}_{\max}(\rhog') :=
\frac{N\lambda_{\max}(\rhog) + (u_{\max}^T h(x))^2}{N+1}
\end{align*}
approximates $\lambda_{\max}(\rhog')$ with error $O(1/N^2)$.
\end{proposition}

Using the estimator $\widehat{\lambda}_{\max}(\rhog')$ from Proposition \ref{prop:lambda_max_update}, we have the following approximation for the change in $\text{VS}_\infty$
\begin{align}
\label{eq:global}
    \Delta_{\text{global}}(x) \approx& -\log \left(\widehat{\lambda}_{\max}(\rhog')\right) + \log \lambda_{\max}(\rhog)
\end{align}
We note that Equation \ref{eq:class} used for the local diversity and Equation \ref{eq:global} used for the global diversity have seemingly similar forms, though Equation \ref{eq:class} is better suited for local diversity measurements because it better handles the variability in class's max eigenvalues. Appendix Section \ref{subsec:local_global} further discusses the differences in these two approaches.

We also scale the global novelty by multiplying by the total dataset size $n$:
\begin{align*}
    S_{\text{GLOBAL-OOD}}(x) =& -n\log \left(\frac{n\lambda_{\max}(\rhog) + \alpha(x)}{n+1}\right) + n\log \lambda_{\max}(\rhog). \notag
\end{align*}

Because subtracting global density does not always yield gains, we introduce a binary coefficient variable $g \in \{0,1\}$, yielding the full VNS detector score:
\begin{align*}
    \text{VNS}(x):= S_{\text{LOCAL-OOD}}(x) - g S_{\text{GLOBAL-OOD}}(x).
\end{align*}
We use $\text{VNS}(x)$ as a continuous OOD score, where larger values indicate more novel samples. To perform OOD detection, this score can be thresholded; throughout our experiments, we choose a threshold that achieves a fixed true positive rate of $95\%$, following standard evaluation practice. Additional details are provided in Section \ref{sec:experiments}.

\textbf{Time and Space Complexity.}
Due to our rank-1 approximation in Equation \ref{eq:class}, VNS has an inference space and time complexity of $O(CD)$, which scales linearly with the number of classes $C$ and the dimension of the data $D$. Logit-based baselines such as MSP \citep{hendrycks2016baseline}, as well as NCI \citep{liu2025detecting}, which leverages neural-collapse structure in the feature space, are still faster. However, VNS is more efficient than Mahalanobis-style methods, which are $O(CD^2)$ \citep{lee2018simple, ren2021simple,  mueller2025mahalanobis++}.

\section{Experiments}
\label{sec:experiments}
In this section, we compare VNS against a diverse set of state-of-the-art OOD detection algorithms. We benchmark all algorithms on three image classification datasets: CIFAR-10, CIFAR-100, and ImageNet-1K, as well as on ResNet, Swin-T, and ViT architectures. Overall, VNS exhibits strong and consistent performance across datasets and models.

\textbf{Datasets and Models.}
Following the OpenOOD benchmark \citep{zhang2023openood}, we use a standardized set of OOD test sets for each ID benchmark. For CIFAR-10 and CIFAR-100, we use six OOD test sets comprising two Near-OOD (hard) and four Far-OOD datasets (easy). 
For CIFAR-10, we use Tiny ImageNet (TIN) \citep{le2015tiny}, MNIST \citep{deng2012mnist}, SVHN \citep{netzer2011reading}, Texture \citep{cimpoi2014describing}, and Places365 \citep{zhou2017places}, and additionally treat CIFAR-100 as an OOD test set.
For CIFAR-100, we use the same five OOD datasets and additionally treat CIFAR-10 as an OOD test set. For both benchmarks, we evaluate using three pretrained ResNet-18 models provided by OpenOOD and report the average across the models. For ImageNet-1K, we use two Near-OOD datasets (SSB-Hard \citep{vaze2021open} and NINCO \citep{bitterwolf2023or}) and three Far-OOD datasets (iNaturalist \citep{van2018inaturalist}, Texture \citep{cimpoi2014describing}, and OpenImage-O \citep{wang2022vim}). We evaluate pretrained ResNet-50 \citep{he2016deep}, ViT-B/16 \citep{dosovitskiy2020image}, and Swin-T \citep{liu2021swin} checkpoints, using 2048-dimensional penultimate-layer features from ResNet-50 and 768-dimensional penultimate-layer features from ViT-B/16 and Swin-T.

\begin{table*}[t]
\centering
\setlength{\tabcolsep}{4.5pt}
\renewcommand{\arraystretch}{1.01}
\resizebox{\textwidth}{!}{
\begin{tabular}{l|cc|cccc|c|cc|cccc|c}
\hline
& \multicolumn{7}{c|}{\textbf{CIFAR-10 OpenOOD Benchmark}} 
& \multicolumn{7}{c}{\textbf{CIFAR-100 OpenOOD Benchmark}} \\
\multirow{2}{*}{\textbf{Method}} &
\multicolumn{2}{c|}{\textbf{Near-OOD}} &
\multicolumn{4}{c|}{\textbf{Far-OOD}} &
\multirow{2}{*}{\textbf{AVG}} &
\multicolumn{2}{c|}{\textbf{Near-OOD}} &
\multicolumn{4}{c|}{\textbf{Far-OOD}} &
\multirow{2}{*}{\textbf{AVG}} \\
& \textbf{\scriptsize CIFAR-100} & \textbf{\scriptsize TIN} & 
  \textbf{\scriptsize MNIST} & \textbf{\scriptsize SVHN} & 
  \textbf{\scriptsize Texture} &
  \textbf{\scriptsize Place365} & &
  \textbf{\scriptsize CIFAR-10} & \textbf{\scriptsize TIN} & 
  \textbf{\scriptsize MNIST} & \textbf{\scriptsize SVHN} & 
  \textbf{\scriptsize Texture} & 
  \textbf{\scriptsize Place365} &\\
\hline

\multicolumn{15}{c}{\textbf{Evaluation under FPR} ($\downarrow$)} \\
MSP & 53.10 & 43.26 & 23.64 & 25.81 & 34.96 & 42.47 & 37.04
            & \second{58.90} & 50.70 & 57.24 & 59.07 & 61.88 & 56.62 & 57.40\\
ASH & 87.31 & 86.29 & 70.00 & 83.64 & 84.60 & 77.86 & 81.62
            & 68.07 & 63.37 & 66.60 & \second{45.97} & 61.29 & 62.94 & 61.37\\
SCALE & 81.78 & 79.12 & 48.69 & 70.54 & 80.39 & 70.51 & 71.84
            & 59.11 & 52.24 & 51.64 & 49.27 & 58.44 & 56.98 & 54.61\\
ADASCALE & 78.93 & 74.89 & 42.20 & 59.97 & 74.75 & 65.90 & 66.11
            & 59.27 & 51.94 & 51.25 & \third{47.41} & 58.39 & 56.41 & 54.11\\
GEN & 58.77 & 48.57 & 23.00 & 28.14 & 40.73 & 47.06 & 41.05
            & \best{58.87} & 49.97 & 53.93 & 55.45 & 61.22 & 56.25 & 55.95\\
IODIN & 61.32 & 50.20 & 25.86 & 31.80 & 43.36 & 49.19 & 43.64
            & \third{59.08} & 51.57 & 52.91 & 54.07 & 62.07 & 57.46 & 56.19\\
KPCA & 41.07 & 32.83 & \third{19.79} & \third{22.68} & \best{22.96} & 30.76 & 28.35
            & 73.54 & 57.01 & \best{45.43} & \best{40.09} & \second{49.11} & 61.09 & 54.38\\
FDBD & \third{39.60} & \third{31.04} & \second{19.33} & 22.89 & 24.28 & \second{29.12} & \third{27.71}
            & 63.89 & \best{47.89} & 51.35 & 53.80 & 53.65 & 57.16 & 54.62\\
NCI & 52.47 & 42.92 & 28.93 & 31.71 & 27.58 & 35.64 & 36.54
            & 63.59 & \second{48.59} & 51.14 & 48.36 & \best{47.76} & \best{53.93} & \best{52.23}\\
KNN & \best{37.62} & \second{30.38} & 20.04 & \second{22.62} & \third{24.06} & \third{30.38} & \second{27.52}
            & 72.81 & 49.66 & \second{48.57} & 51.75 & 53.56 & 60.70 & 56.18\\
RMDS++ & 41.11 & 31.27 & 21.99 & 23.60 & 24.60 & \best{28.64} & 28.53
            & 61.43 & 48.90 & 79.37 & 52.49 & \third{52.40} & \third{54.78} & \third{53.82}\\
\textbf{VNS} & \second{39.48} & \best{30.27} & \best{16.78} & \best{19.68} & \second{23.44} & 31.74 & \best{26.90}
            & 60.34 & \third{48.82} & \third{50.00} & 49.32 & 52.75 & \second{54.04} & \second{52.55}\\
\hline

\multicolumn{15}{c}{\textbf{Evaluation under AUROC} ($\uparrow$)} \\
MSP & 87.19 & 88.87 & 92.63 & 91.46 & 89.89 & 88.92 & 89.83
            & 78.47 & 82.07 & 76.09 & 78.42 & 77.32 & 79.23 & 78.60\\
ASH & 74.10 & 76.44 & 83.16 & 73.45 & 77.45 & 79.89 & 77.42
            & 76.47 & 79.92 & 77.23 & \second{85.60} & 80.72 & 78.76 & 79.78\\
SCALE & 81.27 & 83.84 & 90.58 & 84.63 & 83.94 & 86.41 & 85.11
            & \second{79.26} & 82.71 & 80.27 & 84.45 & 80.50 & 80.47 & 81.28\\
ADASCALE & 82.69 & 85.16 & 91.57 & 87.32 & 85.61 & 87.49 & 86.64
            & 79.22 & 82.82 & 80.39 & \third{85.27} & 81.01 & 80.68 & 81.57\\
GEN & 87.21 & 89.20 & 93.83 & 91.97 & 90.14 & 89.46 & 90.30
            & \best{79.38} & 83.25 & 78.29 & 81.41 & 78.74 & 80.28 & 80.23\\
IODIN & 86.87 & 88.96 & 93.45 & 91.55 & 89.78 & 89.16 & 89.96
            & \third{79.24} & 82.96 & 78.89 & 81.56 & 78.48 & 79.83 & 80.16\\
KPCA & 89.41 & 91.28 & \third{94.46} & \second{92.83} & \second{93.77} & \second{92.05} & \second{92.30}
            & 71.22 & 75.71 & \best{84.64} & \best{89.12} & \best{86.17} & 76.30 & 80.53\\
FDBD & \third{89.56} & \second{91.60} & \second{94.71} & \third{92.80} & 93.13 & \third{92.01} & \second{92.30}
            & 78.35 & \second{83.97} & 79.05 & 80.48 & 81.18 & 79.85 & 80.48\\
NCI & 87.84 & 89.50 & 92.08 & 90.67 & 91.97 & 90.36 & 90.43
            & 78.31 & \third{83.55} & 79.89 & 83.01 & \second{83.75} & \third{80.86} & 81.56\\
KNN & \second{89.73} & \third{91.56} & 94.26 & 92.67 & \third{93.16} & 91.77 & \third{92.19}
            & 77.02 & 83.34 & \second{82.36} & 84.15 & \third{83.66} & 79.43 & \second{81.66}\\
RMDS++ & 88.83 & 90.68 & 92.99 & 92.61 & 92.15 & 91.36 & 91.44
            & 78.08 & 82.72 & 79.37 & 83.95 & 82.56 & \best{83.01} & \third{81.62}\\
\textbf{VNS} & \best{89.97} & \best{92.39} & \best{95.94} & \best{94.43} & \best{93.90} & \best{92.17} & \best{93.13}
            & 78.70 & \best{84.09} & \third{80.99} & 84.47 & 83.06 & \second{82.15} & \best{82.29}\\

\hline
\end{tabular}
}
\caption{VNS achieves state-of-the-art performance on CIFAR-10 and is competitive on CIFAR-100 OpenOOD benchmarks. \best{Orange}, \second{blue}, and \third{purple} mark the best, second-best, and third-best results.}
\label{tab:cifar10_cifar100}
\end{table*}

\textbf{Baselines.}
We compare against $11$ OOD detection algorithms: MSP \citep{hendrycks2016baseline}, ASH \citep{djurisic2022extremely}, SCALE \citep{xu2023scaling}, ADASCALE \citep{regmi2025adascale}, GEN \citep{liu2023gen}, IODIN \citep{regmi2024image}, FDBD \citep{liu2023fast}, NCI \citep{liu2025detecting}, KNN \citep{sun2022out}, KPCA \citep{fang2024kernel}, and
RMDS++ \citep{mueller2025mahalanobis++}. These baselines span a range of widely used post-hoc OOD paradigms: activation shaping (ASH, SCALE, ADASCALE, and IODIN), logit confidences (MSP, GEN), and feature geometry (KPCA, FDBD, NCI, KNN, and RMDS++). Description of each algorithm, and how they compare to VNS is in Appendix Section \ref{sec:baseline}. 

\textbf{Hyperparameter Selection.} 
MSP, FDBD, and RMDS++ are all hyperparameter-free methods and do not require tuning. For all other methods, we follow prior work \citep{liu2025detecting} and the OpenOOD benchmark to automatically select hyperparameters. Specifically, we construct a validation set consisting of ID samples from the training data mixed with synthetic OOD Gaussian noise images, where each pixel is sampled independently from $\mathcal{N}(0,1)$. This protocol provides a proxy for selecting hyperparameters without access to real OOD data.

VNS introduces three hyperparameters: $K$, the number of classes in Equation \ref{eq:aggregate}, $\gamma$, the probability exponent in Equation \ref{eq:aggregate}, and $g$, a binary variable for if the global density correction should be applied. For $K$, we sweep over $\{|C|/100, |C|/40, |C|/20, |C|/10\}$, for $\gamma$, we search over $\{0.1, 0.5, 1., 2.\}$ and for $g$, we search over $\{0,1\}$. We use the hyperparameter triplet $(\gamma, K, g)$ with the highest AUROC on the validation set.

\textbf{Evaluation Metrics.}
We mix each OOD test set with the ID test set, and ask each postprocessor to compute an ID confidence score. Using the scores computed from each method, we compute (1) False Positive Rate under a true positive rate of 95\% (FPR@95) and (2) Area under the ROC curve (AUROC), which measures the probability that a randomly chosen ID sample is deemed less novel than a randomly chosen OOD sample. 

\begin{table*}[!ht]
\centering
\setlength{\tabcolsep}{4.5pt}
\begin{tabular}{l|cc|ccc|c}
\hline
& \multicolumn{6}{c}{\shortstack{\textbf{ImageNet-1K OpenOOD Benchmark}\\
                \textbf{\small (Average Across Architectures)}}} \\
\multirow{2}{*}{\textbf{Method}} &
\multicolumn{2}{c|}{\textbf{Near-OOD}} &
\multicolumn{3}{c|}{\textbf{Far-OOD}} &
\multirow{2}{*}{\textbf{AVG}} \\
& \textbf{\scriptsize SSB-Hard} & \textbf{\scriptsize Ninco} & 
  \textbf{\scriptsize iNaturalist} & \textbf{\scriptsize Textures} & 
  \textbf{\scriptsize OpenImage-O} & {}\\
\hline
\multicolumn{7}{c}{\textbf{Evaluation under FPR} ($\downarrow$)} \\
MSP & 80.60 & 65.12 & 41.06 & 59.41 & 51.87 & 59.61 \\
ASH & 87.62 & 80.91 & 68.45 & 69.98 & 72.57 & 75.91 \\
SCALE & 83.00 & 76.89 & 57.00 & 62.63 & 68.53 & 69.61 \\
ADASCALE & \second{76.43} & 58.96 & 25.30 & 38.64 & \third{33.61} & \third{46.59} \\
GEN & \third{79.04} & \third{54.08} & \third{23.18} & 43.03 & 34.23 & 46.71 \\
IODIN & 81.14 & 65.78 & 36.63 & 54.59 & 47.65 & 57.16 \\
KPCA & 95.40 & 83.98 & 55.79 & 41.97 & 69.01 & 69.23 \\
FDBD & 84.75 & 58.11 & 27.91 & 38.77 & 35.47 & 49.00 \\
NCI & 82.99 & 57.54 & 25.64 & \third{36.70} & 37.06 & 47.99 \\
KNN & 88.51 & 62.36 & 41.06 & \best{27.50} & 47.16 & 53.32 \\
RMDS++ & 83.19 & \second{49.79} & \second{18.64} & 41.62 & \second{32.24} & \second{45.10} \\
\textbf{VNS} & \best{76.12} & \best{47.13} & \best{12.88} & \second{30.26} & \best{28.58} & \best{38.99} \\
\hline
\multicolumn{7}{c}{\textbf{Evaluation under AUROC} ($\uparrow$)} \\
MSP & 70.94 & 79.93 & 88.82 & 83.59 & 85.16 & 81.69 \\
ASH & 57.55 & 61.01 & 64.77 & 62.36 & 64.80 & 62.10 \\
SCALE & 66.15 & 73.06 & 83.44 & 81.68 & 79.22 & 76.71 \\
ADASCALE & \third{71.66} & 82.56 & 93.00 & 90.53 & \second{90.98} & \third{85.74} \\
GEN & 71.62 & \third{83.13} & \third{93.40} & 88.66 & 90.03 & 85.37 \\
IODIN & 71.07 & 80.13 & 89.79 & 85.56 & 86.61 & 82.64 \\
KPCA & 44.41 & 62.44 & 79.04 & 86.79 & 72.53 & 69.04 \\
FDBD & 66.65 & 80.81 & 91.53 & 89.95 & 89.60 & 83.71 \\
NCI & 68.83 & 82.15 & 93.00 & \third{90.79} & 90.15 & 84.98 \\
KNN & 57.45 & 76.45 & 84.38 & \best{93.69} & 83.61 & 79.12 \\
RMDS++ & \second{72.59} & \second{86.09} & \second{95.30} & 89.26 & \third{90.91} & \second{86.83} \\
\textbf{VNS} & \best{75.17} & \best{86.16} & \best{96.71} & \second{91.96} & \best{92.57} & \best{88.51} \\
\hline
\end{tabular}
\caption{VNS achieves state-of-the-art performance on the OpenOOD ImageNet-1K benchmark. Results are averaged across $3$ models (ViT-B/16, ResNet-50, Swin-T). \best{Orange} marks best, \second{blue} marks 2nd, and \third{purple} marks 3rd.}
\label{tab:imagenet_openood_avg}
\end{table*}

\begin{table}[!ht]
\centering
\small
\setlength{\tabcolsep}{5pt}
\begin{tabular}{l|cccccc}
\textbf{Model} & GEN & NCI & ADASCALE & RMDS++ & \textbf{VNS} \\ 
\midrule
 Swin-T Latency
    &  1.19    & 1.19 &  4.72 & 18.54  & 1.48 \\
ResNet-50 Latency
    & 0.59 & 0.59  & 2.50 & 64.66 & 0.78 \\
ViT-B/16 Latency & 2.20 & 2.20 & 9.85 & 14.19 & 2.39\\
\midrule 
\end{tabular}%
\caption{Latency (ms per image) comparison of top-performing post-hoc OOD detectors on ImageNet-1K models. Latency includes model inference, with optimal batch sizes picked for each method. VNS has latency comparable to logit-only methods while outperforming them on the tested benchmarks. Measurements taken on single NVIDIA A6000 GPU.}
\label{tab:latency}
\end{table}

\subsection{OOD Detection Performance Results}
\textbf{CIFAR-10 and CIFAR-100.}
On CIFAR-10, VNS achieves the best average performance across OOD test sets for both FPR@95 and AUROC (Table \ref{tab:cifar10_cifar100}), followed by FDBD and KNN. On CIFAR-100, VNS and RMDS++ \citep{mueller2025mahalanobis++} are the only methods among the top performers on both metrics, with VNS reducing FPR@95 by 2\% relative to RMDS++.

\textbf{ImageNet-1K.}
Table \ref{tab:imagenet_openood_avg} summarizes ImageNet-1K OOD detection results. We average across all tested architectures, with results for individual models shown in Tables \ref{tab:imagenet_vitb16}, \ref{tab:imagenet_resnet}, and \ref{tab:imagenet_swint}. On average, VNS reduces the FPR@95 by $13\%$ relative to the second-best performing algorithm. We note that on individual architectures, other algorithms may perform better. For example, the activation shaping methods SCALE \citep{xu2023scaling} and ADASCALE \citep{regmi2025adascale} perform better than VNS on the ResNet architecture (Table \ref{tab:imagenet_resnet}). However, these gains do not consistently transfer across architectures. \cite{xu2023scaling} argue for the effectiveness of scaling activations under the assumption that the mean of ID activations is higher than that of OOD activations. This claim does not necessarily hold for transformer-based architectures, whereas VNS provides architecturally-robust results.

Table~\ref{tab:latency} shows that VNS is computationally efficient despite using representation-level information. 
Across tested architectures, VNS is orders of magnitude faster than the second-most accurate method RMDS++, and only slightly slower than the fastest detectors GEN and NCI. Thus, VNS retains the accuracy benefits of representation-based OOD detection while avoiding the large test-time overhead of feature-search methods. The efficiency of VNS is possible due to its rank-1 approximation. We show in Section \ref{sec:hyperparam} that using only rank-1 preserves accuracy compared to larger ranks. Additional hyperparameter ablations are provided in Appendix Section \ref{sec:hyperparam}.

\begin{table}[t]
\centering
\setlength{\tabcolsep}{4pt}
\begin{tabular}{ll|cccc}
\toprule
\multicolumn{2}{c|}{\textbf{Benchmark}} 
& \multicolumn{4}{c}{\textbf{Method}} \\
\cmidrule(lr){1-2}
\cmidrule(lr){3-6}
\textbf{Dataset} & \textbf{Architecture}
& GEN+ReAct
& KPCA+GEN
& Hybrid RMDS++
& \textbf{VNS} \\
\midrule
CIFAR-10 & ResNet-18
& 88.59 & \second{92.51} & 91.21 & \best{93.13} \\

CIFAR-100 & ResNet-18
& 80.47 & \best{82.80} & 81.55 & \second{82.29} \\

ImageNet-1K & ResNet-50
& 87.07 & 87.43 & \second{88.10} & \best{89.26} \\

ImageNet-1K & ViT-B/16
& 85.19 & 84.74 & \best{88.35} & \second{87.48} \\

ImageNet-1K & Swin-T
& 85.73 & 85.43 & \second{88.54} & \best{88.79} \\
\midrule 
\end{tabular}
\caption{Average AUROC of hybrid OOD detectors on all tested OpenOOD benchmarks. \best{Orange} and \second{blue} mark the best and second-best methods in each row.}
\label{tab:hybrid}
\end{table}
\subsection{Comparison of Hybrid Approaches}
VNS provides a hybrid approach of combining predictive uncertainty from logits with a class-conditional novelty score computed in feature space. As many other baselines only leverage one set of information, we evaluate VNS against an additional set of hybrid detectors.

One baseline is to combine RMDS++ with our probability-weighting scheme. We follow \citet{mueller2025mahalanobis++} and compute a Mahalanobis distance between a normalized test sample to the class mean. The novelty score function then becomes
\begin{align*}
    s_{\text{Maha}} (x) = \sum_{i \in \mathcal{T}_K(x)} p_c(x)^\gamma d_{\text{Maha}}(x, \mu_c) - g d_{\text{Maha}}(x, \mu),
\end{align*}
where $d_{\text{Maha}}$ denotes the Mahalanobis distance. We also compare VNS against a hybrid of KPCA and the logit-only method GEN, multiplying the scores from the two methods, and a hybrid of GEN and ReAct \citep{sun2021react}. 
Results are shown in Table \ref{tab:hybrid}. The Hybrid-RMDS++ approach provides the most competitive performance to VNS, although there is a larger performance gap on the CIFAR-10/100 benchmarks and RMDS++ can be over $80\times$ slower than VNS (Table \ref{tab:latency}).

\begin{figure}[t]
    \centering
    \includegraphics[width=0.72\linewidth, trim=0.7cm 0.5cm 0 0 ]{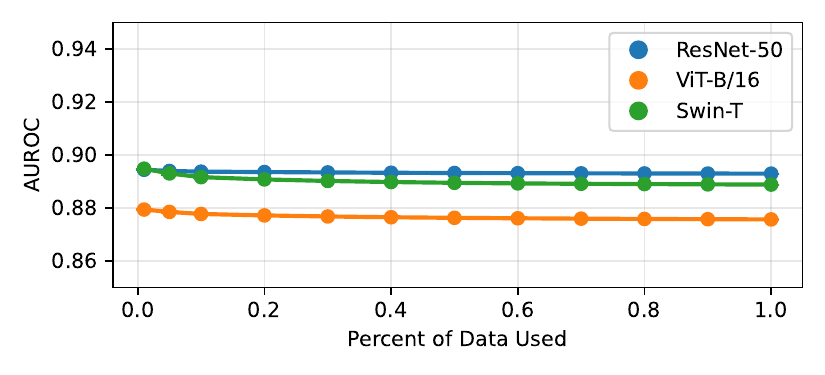}
    \caption{VNS provides accurate OOD detection even with access to only $1\%$ of training data. AUROC is shown as a function of the percent of the training data used. Results across three ImageNet-1K models are displayed, averaged across three reproductions.}
    \label{fig:lessdata}
\end{figure}

\subsection{Data-efficient OOD Detection}
A major advantage of activation shaping and logit-based methods over other approaches is that they do not require access to a model's training data. This is useful in contexts where the training data is not available, or is too expensive to store. While VNS cannot operate without access to the training data, we show that VNS can operate effectively with limited access to the training data. We sample varying percentages of samples from each training ImageNet class and measure VNS's performance after being fit on this subset, using the same hyperparameters throughout. In Figure \ref{fig:lessdata}, we see that even at tiny fractions, VNS provides accurate OOD detection. With $1\%$ of the data, VNS achieves an average AUROC of $88.96 \pm 0.72$ across models, compared to $88.59 \pm 0.74$ when given all of the data. VNS's robustness to dataset size is likely due to two factors: (1) VNS incorporates logit information, which is unaffected by dataset size and (2) VNS only utilizes the largest eigenvalue from each class, which can be reliably estimated from small sample sizes. 

\glsresetall

\section{Conclusion}
This work introduces the Vendi Novelty Score (VNS), a post-hoc OOD detector that leverages the Vendi Scores to quantify novelty from a diversity perspective. VNS scores a test sample by measuring its class-conditional novelty using an efficient rank-1 approximation of the Vendi Score of order 2, aggregates these signals via a probability-weighted top-K scheme, and applies a data-dependent global background correction to improve robustness. Across multiple image classification benchmarks and architectures, VNS often achieves state-of-the-art detection performance. On ImageNet-1K, VNS reduces average FPR@95 by 13\% relative to the strongest tested baseline, and remains effective when computed using only 1\% of the training set.

\textbf{Limitations.} VNS introduces three tunable hyperparameters, which we select using an automatically generated validation set of random noise images. However, this validation set may not be representative of the samples encountered during deployment. In addition, for scalability, VNS relies on the cosine kernel, whereas alternative kernels may improve results. 

\textbf{Broader Impact.}
This work improves OOD detection, helping ML systems identify inputs beyond their competence and trigger safer responses such as abstention, human escalation, or fallback procedures. However, they are not complete safety solutions and can fail under adversarial or shifting conditions. OOD detectors should therefore be used within broader safety protocols with monitoring, human oversight, and evaluation across diverse environments.
\section{Acknowledgements}
We thank the members of Vertaix for their comments. Amey Pasarkar is funded by the NSF
GRFP fellowship. Adji Bousso Dieng is supported by the NSF, OAC \#2118201.

\bibliographystyle{apa}
\bibliography{main}


\section{Additional Algorithm Details for VNS}
\subsection{Scaling Novelty Scores to Account for Dataset Size}
\label{subsec:scaling}
When computing the local and global novelty scores in Equation \ref{eq:class} and Equation \ref{eq:global}, we scale by the number of samples. This removes the effect of dataset size because the increment $\Delta$ scales according to $\frac{1}{N}$. We show this is true for both the local $\Delta_c$ in \ref{eq:class} and $\Delta_{\text{global}}$ in Equation \ref{eq:global}. 

For the class-conditional novelty $\Delta_c$, we can use a Taylor expansion around the terms in Equation \ref{eq:class} to show the scaling behavior of the score.
\begin{align*}
    \Delta_c(x) &= -\log \left( \frac{N_c^2 \lambda_{c}^2 + 2N_c \lambda_{c}\alpha_{c}(x) + 1}{(N_c+1)^2} \right) + \log(\lambda_c^2) \\
    &= -\log \left( \frac{N_c^2 + 2N_c\alpha(x)/\lambda_c + 1/\lambda_c^2}{(N_c+1)^2}\right) \\
    &= 2\log(1 + \frac{1}{N_c}) - \log \left(1+\frac{2\alpha(x)}{N_c\lambda_c} + \frac{1}{N_c^2 \lambda^2}\right) \\ 
    &= 2\left(\frac{1}{N_c} - \frac{1}{2N_c^2} + O(\frac{1}{N_c^3})\right) - \left(\frac{2\alpha(x)}{\lambda_c N_c} + (\frac{1}{\lambda^2}-\frac{2\alpha(x)^2}{\lambda_c^2}) \frac{1}{N_c^2} + O(\frac{1}{N_c^3})\right) \\
    &= (2-\frac{2\alpha(x)}{\lambda_c})\frac{1}{N_c} + O(\frac{1}{N_c^2}),
\end{align*}
where we used the two Taylor expansions $\log(1+u) = u - \frac{u^2}{2} + O(u^3)$ and $\log(1+Au+Bu^2) = Au + (B-\frac{A^2}{2})u^2 + O(u^3)$. With the above, we have
\begin{align*}
    N_c\Delta_c(x) = 2-\frac{2\alpha(x)}{\lambda_c} + O(\frac{1}{n_c}),
\end{align*}
which is independent of the dataset size at the leading order. 

For the global novelty, we again use a Taylor expansion around the terms in Equation \ref{eq:global} to show the scaling behavior of the score. 
\begin{align*}
    \Delta_{\text{global}} &= -\log \left(\frac{N\lambda_{\max}(\rho) + \alpha(x)}{N+1} \right) + \log \lambda_{\max}(\rho) \\
    &= \log \left(1 + \frac{1}{N}\right) - \log \left(1+\frac{\alpha(x)}{N\lambda}\right) \\
    &= \frac{1}{N} - \frac{\alpha(x)}{N\lambda} + O(\frac{1}{N^2})
\end{align*}
Therefore, scaling by a factor of $N$ gives us a score independent of the dataset size at the leading order.
\begin{align*}
    N \Delta_{\text{global}} &= 1 - \frac{\alpha(x)}{\lambda} + O(\frac{1}{N})
\end{align*}

\subsection{Proof of Proposition \ref{prop:lambda_max_update}}
\label{subsec:rayleigh}
As defined in Eq. \ref{eq:trace}, the log-VS of order $\infty$ is given by
\begin{align*}
    \log \text{VS}_{\infty}(\rhog) = -\log \lambda_{\max}(\rhog), 
\end{align*}
where $\lambda_{\max} (\rhog)$ is the largest eigenvalue of $\rhog = X^T X/N \in \mathbb{R}^{D \times D}$. Let $u_{\max}$ be the corresponding eigenvector to $\lambda_{\max} (\rhog)$.
Adding a new sample $x$ to $\rhog$ gives an updated density matrix $\rhog' = \frac{1}{N+1}(N\rhog + h(x) h(x)^T)$. The top eigenvalue of $\rho'$ satisfies the Rayleigh principle 
\begin{align*}
    \lambda_{\max} (\rhog') = \max_{||v||=1} v^T \rhog' v.
\end{align*}
Because the old top eigenvector has $||u_{\max}||=1$, we can use it as a lower bound:
\begin{align*}
     \lambda_{\max} (\rhog') &\geq u_{\max}^T \rho' u_{\max} \\
     &= u^T_{\max} \left(\frac{n\rhog + h(x) h(x)^T}{N+1}\right) u_1 \\
     &= \frac{Nu_{\max}^T \rhog u_{\max} + u_{\max}^T h(x) h(x)^T u_{\max}}{N+1} \\
     &= \frac{N\lambda_{\max}(\rhog) + \alpha(x)}{N+1},
\end{align*}
We plug in the above expression into Eq. \ref{eq:global_def} to reach Eq. \ref{eq:global}. 

We can use a first-order perturbation expansion to derive the accuracy of our approximation. Let $\delta = \rhog' - \rhog = \frac{1}{N+1}\left(h(x) h(x)^T -\rhog \right)$. Then,
\begin{align*}
    \lambda_{\max}(\rhog') &= \lambda_{\max}(\rhog) + u_{\max}^T \delta u_{\max} + O(||\delta||^2) \\
    &= \lambda_{\max}(\rhog) + \frac{(u_{\max}^T h(x))^2 - \lambda_{\max}(\rhog)}{N+1} + O(\frac{1}{N^2}) \\
    &= \frac{N\lambda_{\max}(\rhog) + (u_{\max}^T h(x))^2}{N+1} + O(\frac{1}{N^2}).
\end{align*}
So, our approximation to the max eigenvalue is accurate to the order of $O(1/N^2)$.

\subsection{Comparison of Local and Global Novelty Scores}
\label{subsec:local_global}
Here we compare Equation \ref{eq:class} against Equation \ref{eq:global}. We show that despite their reliance on only the maximum eigenvector and eigenvalue, our $\text{VS}_2$ rank-1 approximation for measuring local diversity is better suited than our first-order $\text{VS}_\infty$ approximation, while $\text{VS}_\infty$ provides an accurate global correction.

In our rank-$1$ approximation in Equation \ref{eq:class}, the data-dependent part of $\Delta_c$ can be written as
\begin{align}
\label{eq:local_q2_rank1}
\Delta_c^{(2)}(x)
\;=\;
-\log\!\left(1 + \frac{2\alpha_c(x)}{N_c\lambda_c} + \frac{1}{N_c^2\lambda_c^2}\right) + \text{Constant}
\end{align}
whereas the analogous $q=\infty$ change (under our first-order approximation) has the form
\begin{equation}
\Delta_c^{(\infty)}(x)
\;=\;
-\log\!\left(1 + \frac{\alpha_c(x)}{N_c\lambda_c}\right) + \text{Constant}
\label{eq:local_qinf_rank1} 
\end{equation}
To understand why $q=2$ is preferable locally, we compare derivatives with respect to the quadratic alignment $\alpha_c(x) = (u_c^T h(x))^2$.
Let $b_c := (N_c\lambda_c)^{-1}$. Differentiating~\eqref{eq:local_q2_rank1} gives
\begin{equation}
\left|\frac{\partial}{\partial \alpha}\Delta_c^{(2)}\right|
=
\frac{2b_c}{1 + 2b_c\alpha + b_c^2},
\qquad
\text{and in particular}\qquad
\left|\frac{\partial}{\partial \alpha}\Delta_c^{(2)}\right|_{\alpha=0}
=
\frac{2b_c}{1+b_c^2}.
\label{eq:q2_derivative}
\end{equation}
In contrast, differentiating~\eqref{eq:local_qinf_rank1} yields
\begin{equation}
\left|\frac{\partial}{\partial \alpha}\Delta_c^{(\infty)}\right|
=
\frac{b_c}{1+b_c\alpha},
\qquad
\text{and in particular}\qquad
\left|\frac{\partial}{\partial \alpha}\Delta_c^{(\infty)}\right|_{\alpha=0}
=
b_c.
\label{eq:qinf_derivative}
\end{equation}
The crucial distinction is that at $\alpha=0$, the $q=\infty$ local sensitivity is linear in $b_c=1/(n_c\lambda_c)$ and can
therefore vary substantially across classes due to variation in $\lambda_c$ (and, to a lesser extent, $N_c$),
making the local novelty overly sensitive to eigenvalue estimation error in small classes. By contrast,
the $q=2$ local sensitivity is self-normalized:
\[
\left|\frac{\partial}{\partial \alpha}\Delta_c^{(2)}\right|_{\alpha=0}
=
\frac{2b_c}{1+b_c^2}
\;\le\; 1,
\]
which suppresses both extremes $b_c\ll 1$ and $b_c\gg 1$. Equivalently, $q=2$ yields a bounded and smoother response
as $\lambda_c$ varies, preventing a small subset of classes with atypical $\lambda_c$ from dominating the
probability-weighted aggregation. This robustness is particularly valuable for local scoring because class-level
statistics are estimated from substantially fewer samples than the global statistics.

At the global-level, $\frac{1}{N\lambda}$ is reliably estimated so we do not need the self-normalization behavior of $\text{VS}_{2}$.

\section{Baselines}
\label{sec:baseline}
Here we briefly describe the $11$ baselines, and how they compare in approach to VNS. All baselines were run on a single NVIDIA A6000 GPU with 4 Intel Xeon Gold 5320, 2.20GHz CPUs. All experimental results were generated over approximately $200$ GPU hours.

\paragraph{MSP.} The maximum softmax probability (MSP) detector from \citet{hendrycks2016baseline} uses the maximum class probability prediction as a measure of model confidence. MSP does not incorporate any information about the feature geometry like VNS.

\paragraph{ASH.} The activation shaping (ASH) method from \citet{djurisic2022extremely} prunes the majority of a sample's activations and scales the rest. These perturbed activations are then fed into the Energy score from \citep{liu2020energy}. The time complexity of ASH is $O(D\log D)$. ASH, unlike VNS, does not require access to the training data. However, this approach can effect ID accuracy unless a second forward pass in the model is made.

\paragraph{SCALE.} \citet{xu2023scaling} found that ASH does not require any pruning, and benefits from only scaling the activations. SCALE also has time complexity $O(D\log D)$, but does not affect ID accuracy. SCALE can be used on any model without accessing training data. 

\paragraph{ADASCALE.} \citet{regmi2025adascale} improved on SCALE by noting that using a fixed scaling factor across all samples can be overly rigid. Instead, ADASCALE uses a gradient perturbation method to estimate an ideal, per-sample scaling factor. This approach requires multiple forward passes as well as a backward pass to compute the scaling factor. There are two variants, ADASCALE-A and ADASCALE-L, that differ in how the scaling is applied. We compare against ADASCALE-A in this paper because it provides better performance on the tested benchmarks. 

\paragraph{IODIN.} \citet{regmi2024image} used an input perturbation approach to achieve better ID-OOD logit separation. In particular, they find the regions of an image that are most informative to a model's prediction and apply a perturbation to them before using the Energy score \citep{liu2020energy}. Like ADASCALE, IODIN requires the ability to compute gradients on the model and multiple forward passes for each test sample. 

\paragraph{GEN.} \citet{liu2023gen} introduced the Generalized Entropy score, a logit-only OOD detector. It measures the entropy of the class predictions over the top-K classes, with OOD samples having higher uncertainty. The entropy is given as $s_\text{GEN}(x) = \sum_{i \in \mathcal{T}_K(x)} p_i(x)^\gamma (1-p_i(x))^\gamma$. We incorporate this form into our probability-weighted aggregation to reflect the model uncertainty alongside our class novelty scores. 

\paragraph{KPCA.} Kernel-PCA for OOD detection was proposed by \cite{fang2024kernel} to measure OOD based on the KPCA reconstruction error. The authors found that using a Gaussian kernel on the $\ell_2$ normalized embeddings, and then approximating this kernel with Random Fourier Features (RFFs) provided a scalable and accurate alternative to standard PCA-reconstruction error pipelines. While KPCA and VNS both rely on kernels,
VNS operate in a kernel/feature space and rely on the eigenspectrum of a similarity operator, but there are two primary differences. (1) KPCA constructs a single, global kernel matrix, whereas VNS constructs local, class-conditional kernel matrices (2) KPCA measures reconstruction error, while VNS measures the change in class diversity via eigenvalue concentration. The primary hyperparameters for KPCA are the RFF dimension, the number of principal components, and the gaussian kernel bandwidth. We use the author recommended defaults for these parameters across datasets, as reconstructing the RFFs based on the dimension or the kernel bandwidth would be too expensive to sweep over. 

\paragraph{FDBD.} The Fast Decision Boundary Detector (FDBD) introduced in \citet{liu2023fast} exploits the observation that OOD samples lie closer to decision boundaries than ID samples in the embedding space. The paper provides a hyperparameter-free, fast $O(C+D)$ algorithm for estimating these distances. The authors also show that FDBD is theoretically accurate under the assumptions of neural collapse. 

\paragraph{NCI.} The Neural Collapse Inspired OOD Detector (NCI) directly leverages the properties of neural collapse to identify OOD samples \citet{liu2025detecting}. In particular, the authors leverage $4$ characteristics of Neural Collapse: (1) variability within a class goes to $0$ (2) class means are arranged on an Equiangular Tight frame (3) the last layer linear classifier converges to the class means and (4) the classifier acts as a nearest center classifier. With these characteristics, NCI measures the cosine similarity between the centered feature and the class weights. NCI is also fast: it runs in $O(D)$ additional time. NCI introduces one hyperparameter to incorporate the norm of test samples.

\paragraph{KNN.} The KNN OOD detector from \citet{sun2022out} measures the distance between a test sample and its $k$th nearest neighbor in a normalized feature space. This non-parametric method requires each test sample to be compared against the entire training set. At inference time, VNS does not search over the entire training dataset.

\paragraph{RMDS++.}
Given $\ell_2$-normalized embeddings $x \in \mathbb{R}^D$ with $\|x\|_2 = 1$, RMDS++ measures OOD uncertainty by comparing how well $x$ fits a class-conditional Gaussian model versus the global feature distribution \citep{ren2021simple, mueller2025mahalanobis++}. Let $\mu_c, \Sigma$ denote the mean and covariance of class $c$, and let $\mu_g, \Sigma_g$ denote the mean and covariance computed over the full training set. $\Sigma$ is shared across all classes. RMDS++ uses a \emph{relative} Mahalanobis distance score as
\begin{equation}
    s_{\mathrm{RMDS++}}(x)
    =
    \min_{c \in [C]}
    \left[
        (x-\mu_c)^\top \Sigma^{-1}(x-\mu_c)
        -
        (x-\mu_g)^\top \Sigma_g^{-1}(x-\mu_g)
    \right],
\end{equation}
where smaller values indicate that $x$ is substantially closer to some class distribution than to the global density. \citet{mueller2025mahalanobis++} had found that using normalized embeddings vastly improved detector performance across a wide variety of architectures. 

\begin{table}[t]
\centering
\begin{tabular}{lccc}
\toprule
\textbf{Ablation} & \textbf{ResNet-50} & \textbf{ViT-B/16} & \textbf{Swin-T} \\
\midrule
Original 
    & 89.26 & 87.48 & 88.79 \\
No probability-weighting 
    & 84.98 & 87.28 & 87.30 \\
No global correction ($g=0$) 
    & 89.25 & 87.48 & 86.73 \\
Global correction ($g=1$) 
    & 89.26 & 87.45 & 88.79 \\
$\gamma=1$ 
    & 87.45 & 86.75 & 86.98 \\
$\gamma=0.5$ 
    & 88.59 & 87.01 & 88.79 \\
$\gamma=0.1$ 
    & 89.26 & 87.48 & 87.53 \\
$K=1000$ 
    & 88.25 & 87.47 & 88.56 \\
$K=100$ 
    & 89.26 & 86.72 & 87.94 \\
$K=10$ 
    & 88.38 & 87.48 & \textbf{88.94} \\
$\text{rank}=5$ & \textbf{89.29}  & 87.54 & 88.86  \\
$\text{rank}=10$ & 89.28  & 87.56 & 88.87   \\
$\text{rank}=50$ & 89.28  & \textbf{87.59} & 88.89  \\
\bottomrule
\end{tabular}
\caption{Ablation study of VNS on ImageNet-1K OOD Average AUROC detection across ResNet-50, ViT-B/16, and Swin-T. Best performance in each column is bolded.}
\label{tab:vns_ablation}
\end{table}

\section{VNS Ablation Analysis}
\label{sec:hyperparam}
We perform hyperparameter ablations of VNS on the tested benchmarks to justify the role of each component as well as the choice of our rank-1 approximation. In Table \ref{tab:vns_ablation}, we test how VNS performs if we do not use the probability-weighted aggregation and instead only use the minimum class-conditional score. For other tests, we fix one hyperparameter and do the automatic selection on the validation set for the rest. The three hyperparameters are: $g$, the global correction term, 
$\gamma$, a class-probability weighting exponent used in Equation \ref{eq:aggregate}, and  $K$, the number of classes to consider. Table \ref{tab:vns_ablation} shows that the probability-weighting scheme is important for our performance, which is unsurprising given we only track the principal eigenvector and eigenvalue in each class. Leveraging information from multiple classes helps us compensate for using less class-specific information. For different choices of hyperparameters, we can achieve competitive, and sometimes even better, performance than the selected parameters.

\begin{figure}[t]
    \centering
    \includegraphics[width=0.75\linewidth, trim=0.5cm 0.5cm 0 0 ]{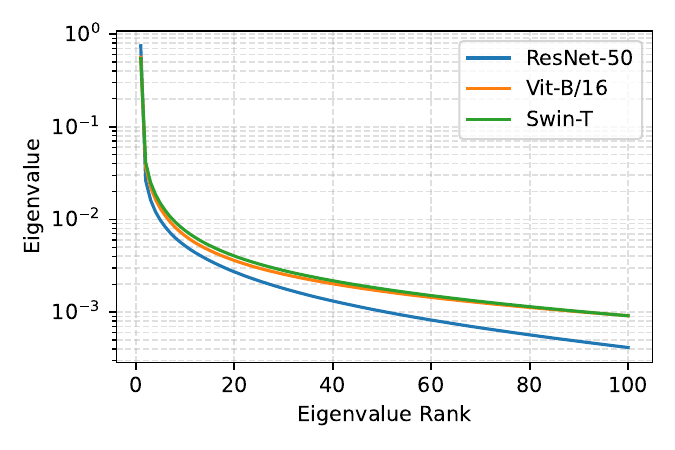}
    \caption{Average eigenspectrum of class-conditional density matrices $\rho_c$ across models is dominated by the leading eigenvalue. The average values of the top-100 eigenvalues of $\rho_c$ are shown (log-scale $y$-axis), across three ImageNet-1K models.}
    \label{fig:eigenspectrum}
\end{figure}

In Table \ref{tab:vns_ablation}, we also include average AUROC for using different choices of rank from Equation \ref{eq:class}. In general, only using a rank-$1$ approximation provides very similar performance to using a rank-$50$ approximation.
In Figure \ref{fig:eigenspectrum}, we observe that class density matrices exhibit a large spectral gap: the leading eigenvalue is on average an order of magnitude larger than the second. Since $\text{VS}_2$ depends on the squared spectrum (Equation \ref{eq:trace}), this gap implies that the rank-1 component captures most of the $\text{VS}_2$-relevant information, whereas smaller components contribute weakly and are more sensitive to noise.

\begin{table*}[t]
\centering

\setlength{\tabcolsep}{4.5pt}
\begin{tabular}{l|cc|ccc|c}
\hline
& \multicolumn{6}{c}{\shortstack{\textbf{ImageNet-1K OpenOOD Benchmark}\\
                \textbf{\small (ViT-B/16)}}} \\
\multirow{2}{*}{\textbf{Method}} &
\multicolumn{2}{c|}{\textbf{Near-OOD}} &
\multicolumn{3}{c|}{\textbf{Far-OOD}} &
\multirow{2}{*}{\textbf{AVG}} \\
& \textbf{\scriptsize SSB-Hard} & \textbf{\scriptsize Ninco} & 
  \textbf{\scriptsize iNaturalist} & \textbf{\scriptsize Textures} & 
  \textbf{\scriptsize OpenImage-O} & {}\\
\hline

\multicolumn{7}{c}{\textbf{Evaluation under FPR} ($\downarrow$)} \\
MSP         & 86.42 & 77.28 & 42.40 & 56.46 & 56.19 & 63.75 \\
ASH         & 93.50 & 95.37 & 97.02 & 98.50 & 94.79 & 95.84
\\
SCALE       & 92.35 & 94.66 & 86.73 & 84.69 & 89.50 &  89.59 \\
ADASCALE    & \third{85.89} & 62.79 & 36.45 & 54.01 & 43.65 & 56.56 \\
GEN         & \second{82.23} & \second{59.33} & \second{22.92} & \third{38.30} & \second{35.47} & \second{47.65} \\
IODIN       & 86.41 & 77.33 & 42.47 & 56.45 & 56.18 & 63.77 \\
KPCA     &   92.00    &  72.32     &     47.94  &  49.20    & 53.66      &  63.02     \\
FDBD        & 87.03 & \third{60.67} & \third{33.14} & 45.07 & \third{39.56} & \third{53.09} \\
NCI         & 86.14 & 61.46 & 34.84 & 48.86 & 41.96 & 54.65 \\
KNN         & 89.04 & 63.90 & 45.62 & \best{21.41} & 50.97 & 54.19 \\
RMDS++      & 84.82 & 45.07 & 13.95 & 35.71 & 27.23 & \second{41.36} \\
\textbf{VNS}    & \best{83.79} & \best{45.31} & \best{16.53} & \second{34.37} & \best{26.42} & \best{41.28} \\
\hline

\multicolumn{7}{c}{\textbf{Evaluation under AUROC} ($\uparrow$)} \\
MSP         & \third{68.94} & 78.11 & 88.19 & 85.06 & 84.86 & 81.03 \\
ASH         & 53.90 & 52.51 & 50.62 & 48.53 & 55.51 & 52.21 \\
SCALE       & 56.46 & 61.33 & 73.69 & 78.88 & 72.54 & 68.58 \\
ADASCALE    & 67.47 & 80.55 & 89.81 & 86.77 & 87.86 & 82.49 \\
GEN         & \second{70.09} & \second{82.51} & \second{93.54} & \third{90.23} & \second{90.27} & \second{85.33} \\
IODIN       & 68.94 & 78.11 & 88.19 & 85.06 & 84.86 & 81.03 \\
KPCA     &   51.62    & 70.10      &  80.36     &  82.08     &  77.78     & 72.39      \\
FDBD        & 65.80 & \third{80.75} & \third{90.46} & 88.50 & \third{89.01} & \third{82.90} \\
NCI         & 66.30 & 80.80 & 90.20 & 87.85 & 88.46 & 82.72 \\
KNN         & 55.64 & 76.06 & 83.28 & \best{96.28} & 83.18 & 78.89 \\
RMDS++      & 73.72 & 88.31 & 96.82 & 89.11 & 92.39 & \best{88.07} \\
\textbf{VNS}    & \best{72.99} & \best{86.23} & \best{95.58} & \second{90.38} & \best{92.23} & \second{87.48} \\
\hline
\end{tabular}
\caption{VNS provides competitive performance with state-of-the-art methods on the OpenOOD ImageNet-1K benchmark with a ViT-B/16 model. RMDS++ achieves similar performance. \best{Orange} marks best, \second{blue} marks 2nd, and \third{purple} marks 3rd.}
\label{tab:imagenet_vitb16}
\end{table*}

\begin{table*}[t]
\centering

\setlength{\tabcolsep}{4.5pt}
\begin{tabular}{l|cc|ccc|c}
\hline
& \multicolumn{6}{c}{\shortstack{\textbf{ImageNet-1K OpenOOD Benchmark}\\
                \textbf{\small (Resnet-50)}}} \\
\multirow{2}{*}{\textbf{Method}} &
\multicolumn{2}{c|}{\textbf{Near-OOD}} &
\multicolumn{3}{c|}{\textbf{Far-OOD}} &
\multirow{2}{*}{\textbf{AVG}} \\
& \textbf{\scriptsize SSB-Hard} & \textbf{\scriptsize Ninco} & 
  \textbf{\scriptsize iNaturalist} & \textbf{\scriptsize Textures} & 
  \textbf{\scriptsize OpenImage-O} & {}\\
\hline

\multicolumn{7}{c}{\textbf{Evaluation under FPR} ($\downarrow$)} \\
MSP         & 74.48 & 56.87 & 43.46 & 60.89 & 50.13 & 57.15\\
ASH         & \third{73.66} & \third{52.99} & 14.10 & 15.28 & 29.19 & 37.04 \\
SCALE       & \second{67.71} & \second{51.81} & \third{9.51} & \second{11.90} & \second{28.17} & \second{33.82} \\
ADASCALE    & \best{57.87} & \best{45.76} & \second{7.60} & \best{10.42} & \best{20.59} & \best{28.45} \\
GEN         & 75.71 & 54.88 & 26.11 & 46.23 & 34.52 & 47.49\\
IODIN       & 76.14 & 58.78 & 30.06 & 46.40 & 37.46 & 49.75 \\
KPCA   & 98.31      &    92.13   &     58.93  &  23.22     &   82.89    &  73.57     \\
FDBD        & 81.00 & 54.28 & 23.23 & 28.96 & 31.85 & 43.86 \\
NCI         & 75.76 & 53.72 & 14.84 & 18.08 & 32.19 & 38.92 \\
KNN         & 89.04 & 63.90 & 45.62 & 21.41 & 50.97 & 54.19 \\
RMDS++      & 80.70 & 53.20 & 29.85 & 48.42 & 39.88 & 50.41 \\
\textbf{VNS}    & 74.55 & 54.16 & \best{6.89} & \third{13.48} & \third{33.63} & \third{36.54} \\
\hline

\multicolumn{7}{c}{\textbf{Evaluation under AUROC} ($\uparrow$)} \\
MSP         & 72.09 & 79.95 & 88.41 & 82.43 & 84.86 & 81.55 \\
ASH         & \third{72.89} & 83.45 & 97.06 & 96.90 & \third{93.26} & 88.71 \\
SCALE       & \second{77.35} & \second{85.37} & \third{98.02} & \second{97.63} & \second{93.95} & \second{90.46} \\
ADASCALE    & \best{81.66} & \best{87.14} & \second{98.31} & \best{97.88} & \best{95.62} & \best{92.12} \\
GEN         & 72.01 & 81.70 & 92.44 & 87.60 & 89.26 & 84.60 \\
IODIN       & 72.49 & 80.57 & 91.33 & 88.35 & 89.21 & 84.39 \\
KPCA     &   40.03    & 58.13      &   82.36   &  95.70    &   73.65    &  69.97     \\
FDBD        & 68.22 & 81.55 & 93.22 & 92.99 & 90.32 & 85.26 \\
NCI         & 72.62 & 83.20 & 96.86 & 96.53 & 92.69 & 88.38 \\
KNN         & 55.64 & 76.06 & 83.28 & 96.28 & 83.18 & 78.89 \\
RMDS++      & 71.64 & 83.52 & 91.99 & 90.46 & 88.73 & 85.27 \\
\textbf{VNS}    & 73.82 & \third{84.32} & \best{98.35} & \third{97.15} & 92.71 & \third{89.27} \\
\hline
\end{tabular}
\caption{VNS ranks third on the OpenOOD ImageNet-1K benchmark with a ResNet-50 model. Activation shaping methods SCALE and ADASCALE perform best. \best{Orange} marks best, \second{blue} marks 2nd, and \third{purple} marks 3rd.}
\label{tab:imagenet_resnet}
\end{table*}

\begin{table*}[t]
\centering
\setlength{\tabcolsep}{4.5pt}
\begin{tabular}{l|cc|ccc|c}
\hline
& \multicolumn{6}{c}{\shortstack{\textbf{ImageNet-1K OpenOOD Benchmark}\\
                \textbf{\small (Swin-T)}}} \\
\multirow{2}{*}{\textbf{Method}} &
\multicolumn{2}{c|}{\textbf{Near-OOD}} &
\multicolumn{3}{c|}{\textbf{Far-OOD}} &
\multirow{2}{*}{\textbf{AVG}} \\
& \textbf{\scriptsize SSB-Hard} & \textbf{\scriptsize Ninco} & 
  \textbf{\scriptsize iNaturalist} & \textbf{\scriptsize Textures} & 
  \textbf{\scriptsize OpenImage-O} & {}\\
\hline

\multicolumn{7}{c}{\textbf{Evaluation under FPR} ($\downarrow$)} \\
MSP         & 80.90 & 61.21 & 37.31 & 60.87 & 49.30 & 57.92 \\
ASH         & 95.70 & 94.38 & 94.24 & 96.16 & 93.74 & 94.84 \\
SCALE       & 88.94 & 84.20 & 74.76 & 91.30 & 87.91 & 85.42 \\
ADASCALE    & 85.54 & 68.33 & 31.84 & 51.49 & 36.59 & 54.76 \\
GEN         & \second{79.19} & \second{48.04} & \third{20.51} & 44.57 & \third{32.69} & \third{45.00} \\
IODIN       & \third{80.88} & 61.24 & 37.35 & 60.93 & 49.30 & 57.94 \\
KPCA     &  95.88    & 87.49      &   60.50   &  53.50    &   70.49    &  73.57     \\
FDBD        & 86.23 & 59.39 & 27.35 & \third{42.29} & 35.00 & 50.05 \\
NCI         & 87.07 & 57.44 & 27.25 & 43.15 & 37.04 & 50.39 \\
KNN         & 87.44 & 59.27 & 31.93 & \best{39.68} & 39.55 & 51.57 \\
RMDS++      & 84.05 & \third{51.10} & \best{12.11} & \second{40.74} & \second{29.60} & \second{43.52} \\
\textbf{VNS}    & \best{70.01} & \best{41.93} & \second{15.22} & 42.94 & \best{25.69} & \best{39.16} \\
\hline

\multicolumn{7}{c}{\textbf{Evaluation under AUROC} ($\uparrow$)} \\
MSP         & 71.78 & 81.72 & 89.86 & 83.27 & 85.77 & 82.48 \\
ASH         & 45.87 & 47.07 & 46.62 & 41.66 & 45.64 & 45.37 \\
SCALE       & 64.64 & 72.47 & 78.60 & 68.52 & 71.18 & 71.08 \\
ADASCALE    & 65.84 & 79.98 & 90.87 & 86.93 & 89.45 & 82.61 \\
GEN         & \second{72.75} & \third{85.18} & \third{94.23} & 88.15 & \third{90.56} & \third{86.17} \\
IODIN       & 71.79 & 81.72 & 89.86 & 83.28 & 85.77 & 82.48 \\
KPCA     &     41.59  &  59.08     &    74.40   &  82.58     &  66.17     &  64.76     \\
FDBD        & 65.93 & 80.12 & 90.91 & \second{88.35} & 89.48 & 82.96 \\
NCI         & 67.58 & 82.44 & 91.95 & 87.99 & 89.29 & 83.85 \\
KNN         & 61.06 & 77.23 & 86.58 & \best{88.51} & 84.47 & 79.57 \\
RMDS++      & \third{72.42} & \second{86.45} & \best{97.08} & 88.21 & \second{91.62} & \second{87.16} \\
\textbf{VNS}    & \best{78.71} & \best{87.93} & \second{96.20} & \third{88.34} & \best{92.76} & \best{88.79} \\
\hline
\end{tabular}
\caption{VNS achieves state-of-the-art performance on the OpenOOD ImageNet-1K benchmark with a Swin-T model. \best{Orange} marks best, \second{blue} marks 2nd, and \third{purple} marks 3rd.}
\label{tab:imagenet_swint}
\end{table*}

\end{document}